\pdfoutput=1
\documentclass[11pt]{article}
\usepackage{acl}
\usepackage{times}
\usepackage{latexsym}
\usepackage[T1]{fontenc}
\usepackage[utf8]{inputenc}
\usepackage{microtype}

\usepackage{booktabs, tabularx}
\usepackage{graphics}
\usepackage{amsmath}
\usepackage{soul,array,multirow,graphicx}
\usepackage{caption}
\usepackage{subcaption}
\usepackage{tikz-dependency}
\usepackage{pgfplots}
\pgfplotsset{compat=1.17}
\usepackage{CJKutf8}
\usepackage{algorithm}
\usepackage{algpseudocode}

\newcommand{\flag}[1]{#1}
\newcommand{\declarelogo}[0]{\includegraphics[height=.02\textwidth]{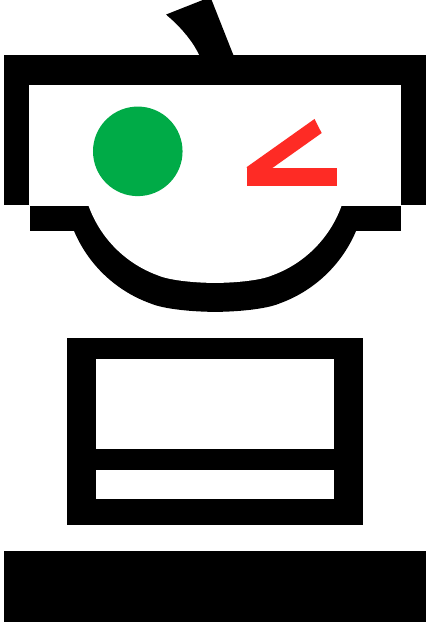}}

\title{RelationPrompt: Leveraging Prompts to Generate Synthetic Data for Zero-Shot Relation Triplet Extraction}
\author{
\textbf{
Yew Ken Chia\thanks{$^{*}$Yew Ken is a student under the Joint PhD Program between Alibaba and SUTD. 
}
~\textsuperscript{\rm 1,${\declarelogo}$}\quad
Lidong Bing\thanks{~~Corresponding author.}~\textsuperscript{\rm 1}\quad
Soujanya Poria\textsuperscript{\rm ${\declarelogo}$}\quad
Luo Si\textsuperscript{\rm 1}}\\
\textsuperscript{\rm 1}DAMO Academy, Alibaba Group~~
\textsuperscript{\rm ${\declarelogo}$} Singapore University of Technology and Design ~~\\
{\tt\{yewken.chia, l.bing, luo.si\}@alibaba-inc.com} \\~~{\tt\{yewken\_chia, sporia\}@sutd.edu.sg}}

\begin{document}
\maketitle

\begin{abstract}
Despite the importance of relation extraction in building and representing knowledge, less research is focused on generalizing to unseen relations types.
We introduce the task setting of Zero-Shot Relation Triplet Extraction (ZeroRTE) to encourage further research in low-resource relation extraction methods. 
Given an input sentence, each extracted triplet consists of the head entity, relation label, and tail entity where the relation label is not seen at the training stage.
To solve ZeroRTE, we propose to synthesize relation examples by prompting language models to generate structured texts.
Concretely, we unify language model prompts and structured text approaches to design a structured prompt template for generating synthetic relation samples when conditioning on relation label prompts (RelationPrompt).
To overcome the limitation for extracting multiple relation triplets in a sentence, we design a novel Triplet Search Decoding method.
Experiments on FewRel and Wiki-ZSL datasets show the efficacy of RelationPrompt for the ZeroRTE task and zero-shot relation classification. 
Our code and data are available at
\href{https://github.com/declare-lab/RelationPrompt}{github.com/declare-lab/RelationPrompt}.

\end{abstract}

\section{Introduction}

Relation extraction aims to predict relationships between entities in unstructured text, which has applications such as knowledge graph construction \cite{lin2015learning} and question answering \cite{xu2016question}.
However, existing approaches often require large datasets of annotated samples which are costly to annotate and have a fixed set of relations.
Currently, less research is focused on the zero-shot setting \cite{wang2019survey} where models need to generalize to unseen relation sets without available annotated samples \cite{wang2019survey}.
Although there are existing zero-shot relation task settings, they do not require extracting the full relation triplets.
The task setting of Zero-Shot Relation Classification\footnote{As relation classification and relation extraction are sometimes interchangeable, we refer to relation classification.}
(ZeroRC) was previously introduced by \citet{chen2021zs} to classify the relation between a given head and tail entity pair for unseen labels. 
However, it is not always practical or realistic to assume that the ground-truth entities are readily available.
Zero-Shot Relation Slot-Filling \cite{levy2017zero} aims to predict the tail entity based on the provided head entity and relation, but also relies on other methods for entity detection. 
Thus, it also faces the challenge of error propagation in practice \cite{zhong-chen-2021-frustratingly}.

\begin{table}[!t]
    \includegraphics[width=1.0\columnwidth]{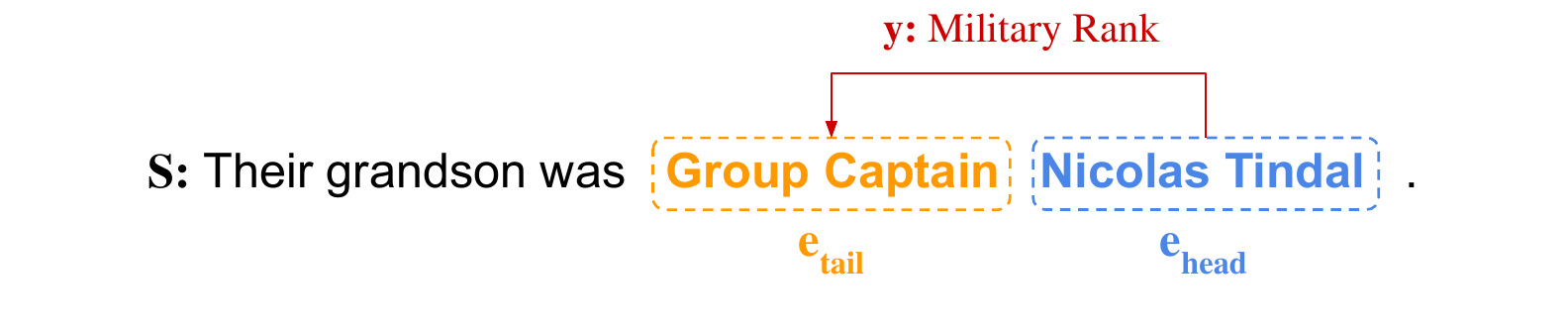}
    
    \centering
    \resizebox{1\columnwidth}{!}{
    \begin{tabular}{llccc}
    \toprule
    & Task Setting & Input & Output & Supervision \\
    \midrule
    & Relation Classification
    & $S, e_{head}, e_{tail}$ & $y$ & Full  \\
    & Zero-Shot Relation Classification
    & $S, e_{head}, e_{tail}$ & $y$ & Zero-Shot  \\
    & Zero-Shot Relation Slot-Filling
    & $S, e_{head}, y$ & $e_{tail}$ & Zero-Shot  \\
    & Relation Triplet Extraction
    & $S$ & $e_{head}, e_{tail}, y$ & Full \\
    & \textbf{Zero-Shot Relation Triplet Extraction}
    & $S$ & $e_{head}, e_{tail}, y$ & Zero-Shot \\

  \bottomrule
    \end{tabular}
    }
   \caption{Comparison of task settings with our proposed Zero-Shot Relation Triplet Extraction (ZeroRTE). 
   To our knowledge, ZeroRTE is the first task to extract full relation triplets in the zero-shot setting.} 
    \label{tab:tasks}
\end{table}

\begin{figure}[!t]
\centering
\includegraphics[width=1.0\columnwidth]{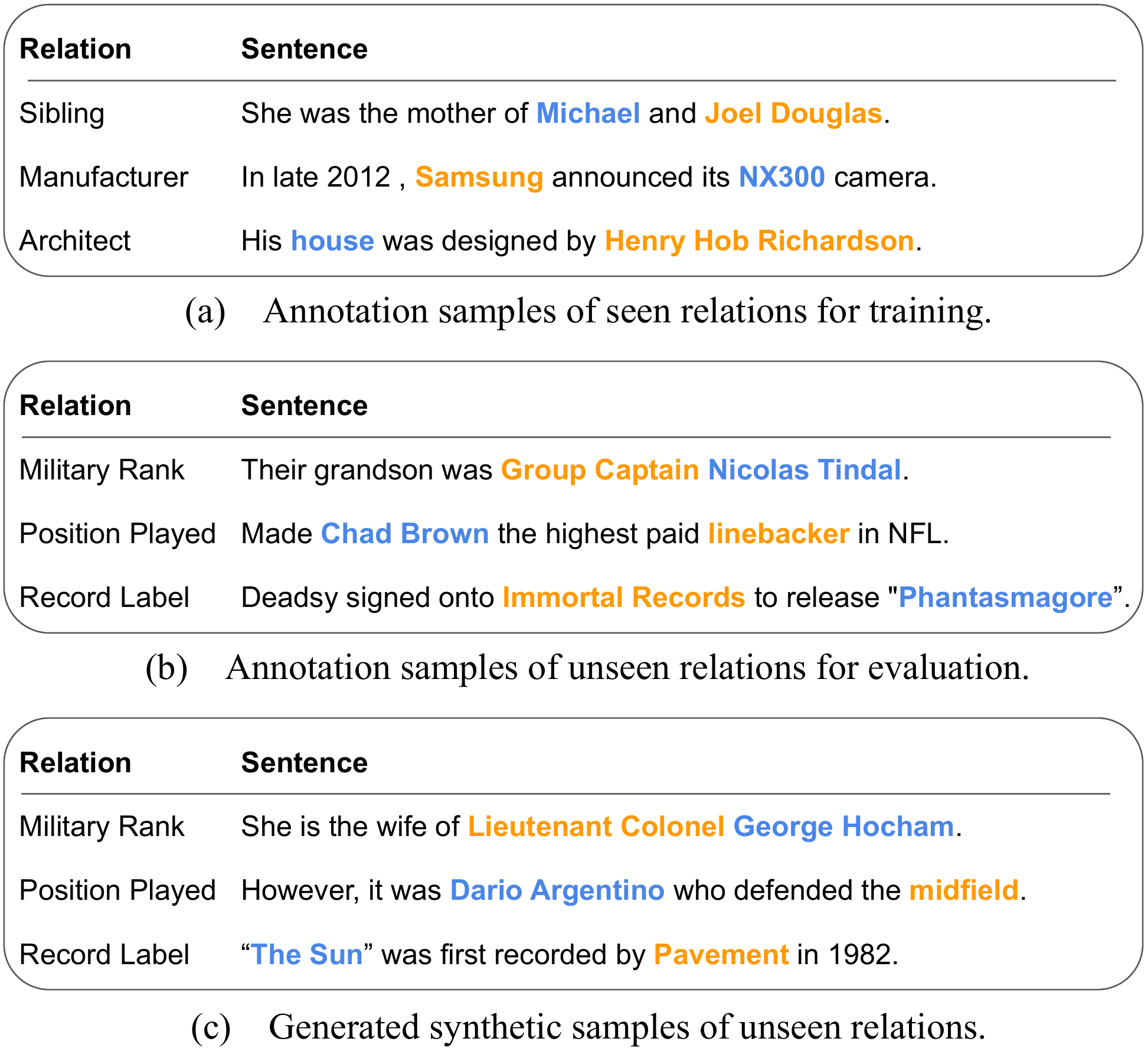}
\caption{
Example relation triplet data for ZeroRTE and our formulation as synthetic sentence generation.
The head and tail entities are shown in blue and orange, respectively.
The ZeroRTE train samples (a) and test samples (b) contain triplets that belong to disjoint relation label sets.
We formulate ZeroRTE as generating synthetic samples (c) for the unseen test relation labels.
The synthetic data can then be used to train another model to extract relation triplets from the test sentences.
We also present more data samples in Appendix \ref{sec:more_samples}.
}
\label{fig:example}
\end{figure}

Hence, we propose a new and challenging task setting called Zero-Shot Relation Triplet Extraction (ZeroRTE).
The goal of ZeroRTE is to extract triplets of the form (head entity, tail entity, relation label) from each sentence despite not having any annotated training samples that contain the test relation labels. 
For a clear comparison between task settings, we provide a summary in Table \ref{tab:tasks}.
To our knowledge, this is the first work to extend the task of Relation Triplet Extraction to the zero-shot setting.
For example in Figure \ref{fig:example}, the training samples may belong to the seen relation set \{Sibling, Manufacturer, Architect\}, while the test samples may belong to the unseen relation set \{Military Rank, Position Played, Record Label\}.
Given the annotated training samples in Figure \ref{fig:example}a, ZeroRTE aims to extract triplets such as (Nicolas Tindal, Military Rank, Captain) in Figure \ref{fig:example}b.

To solve the challenges of data scarcity, there are several existing approaches.
Although distant supervision \cite{ji2017distant} can be used to construct a relation corpus with a many relation types, this approach generally results in lower annotation quality than human annotation. Furthermore, distant supervision remains limited to a fixed set of relation types in the existing knowledge base \cite{smirnova2018relation}.
Another approach is to formulate the task objective such that the label space is unconstrained.
For instance, zero-shot sentence classification can be reframed as entailment \cite{puri2019zero} or embedding similarity \cite{pushp2017train} objectives.
However, the existing formulations 
are designed for sequence classification tasks, which 
cannot be directly applied to 
structured prediction
tasks such as relation triplet extraction. 
A third direction is to leverage pre-trained language models using task-specific prompt templates \cite{liu2021pre} which enables the models to generalize to new tasks with little to no training samples, such as zero-text classification \cite{zhong2021adapting}.
This zero-shot potential is possible by leveraging the semantic information in prompts to query the language comprehension capabilities of pre-trained language models \cite{radford2019language}.

Hence, we propose RelationPrompt which reframes the zero-shot problem as synthetic data generation.
The core concept is to leverage the semantics of relation labels, prompting language models to generate synthetic training samples which can express the desired relations. 
The synthetic data can then be used to train another model to perform the zero-shot task.
This capability is supported by the finding that language models can be prompted to control task-specific aspects of the generated text, such as domain and content \cite{keskar2019ctrl}.
For instance, given the relation label ``Military Rank'' in Figure \ref{fig:example}c, it is reasonable to condition the language model and compose a sentence demonstrating the relationship
that a person has been bestowed with a certain position in the armed forces.
Hence, a possible sentence could be ``She is the wife of Lieutenant Colonel George Hocham.'', 
where the head entity is ``George Hocham'' and the tail entity is ``Lieutenant Colonel''. 
Given generated samples of sufficient quality and diversity, the synthetic dataset can effectively supervise another model to perform ZeroRTE. 

To encode the relation triplet information as text sequences which can be generated by language models, we unify prompt templates with structured text formats \cite{paolini2020structured}.
Structured texts use special markers to encode the structured information which can be easily decoded as triplets. 
However, it is challenging to generate sentences which contain multiple different relation triplets.
Designing a complex structured prompt template to encode multiple triplets may compromise the generation quality as the language model needs to manipulate multiple relations at once.
Hence, we focus on generating single-triplet samples and explore how this limitation can be overcome by the downstream relation extractor model.
Concretely, we propose a method named Triplet Search Decoding which allows the extraction of multiple triplets at prediction time despite training on synthetic samples which contain a single triplet each.
\textbf{Contributions.} In summary, our main contributions include:
(1) We introduce the ZeroRTE task setting which overcomes limitations in prior task settings by extending the Relation Triplet Extraction task to the zero-shot setting. ZeroRTE is released as a publicly available benchmark based on the reorganized FewRel \cite{han2018fewrel} and Wiki-ZSL \cite{chen2021zs} datasets.
(2) In order to make ZeroRTE solvable in a supervised manner, we propose RelationPrompt to generate synthetic relation examples by prompting language models to generate structured texts.
(3) We propose Triplet Search Decoding to overcome the limitation for extracting multiple relation triplets in a sentence.
(4) RelationPrompt surpasses prior ZeroRC methods and baselines on ZeroRTE, setting the bar for future work.
Our analysis shows that the generated samples are reasonable and diverse, hence serving as effective synthetic training data.



\section{RelationPrompt: Methodology}
To extract triplets for unseen relation labels in ZeroRTE, we propose a framework called RelationPrompt which uses relation labels as prompts to generate synthetic relation examples of target unseen labels.
The synthetic data can then be used to supervise any downstream relation extraction model.
Hence, our framework requires two models: a Relation Generator for synthetic relation samples, and a Relation Extractor that will be trained on the synthetic data and used to predict triplets for unseen relations.
In order to represent the relation triplet information to be processed by language models, we design structured prompt templates.
The relation extractor is designed to support both ZeroRTE and ZeroRC tasks.
We further propose Triplet Search Decoding to overcome the challenge of generating relation samples with multiple triplets.

\subsection{Task Formulation}
\flag{
In ZeroRTE, the goal is to learn from the seen dataset $D_s$ and generalize to the unseen dataset $D_u$.
The datasets $D_s$ and $D_u$ are used for training and testing respectively, and are originally split from the full dataset which is defined as $D = (S, T, Y)$ where $S$ denotes the input sentences, $T$ denotes the output triplets and $Y$ denotes the set of relation labels present in $D$. 
The seen and unseen label sets are predefined and denoted as 
$Y_{s} = \{y_{s}^{1}, ..., y_{s}^{n}\}$
and
$Y_{u} = \{y_{u}^{1}, ..., y_{u}^{m}\}$
respectively, where 
$n = |Y_{s}|$
and
$m = |Y_{u}|$
are the size of seen and unseen label sets respectively.
Hence, the label sets of $D_s$ and $D_u$ are disjoint, i.e., $Y_{s} \cap Y_{u} = \emptyset$.
Each data sample contains the input sentence $s \in S$ which corresponds to a list $t \in T$ which can contain one or more output triplets.
A relation triplet is defined as 
$(e_{head}, e_{tail}, y)$ which denotes the head entity, tail entity and relation label respectively.
To solve ZeroRTE, we formulate the following algorithm:
}

\begin{algorithm}
\caption{RelationPrompt: Prompting language models to generate synthetic data for ZeroRTE.}
\label{alg:cap}
\textbf{Define:} \\
Dataset $D = (\textrm{Sentences } S, \textrm{Triplets } T, \textrm{Labels } Y)$
\begin{algorithmic}[1]
\Require
Train Dataset $D_s$,
Test Dataset $D_u$,
Relation Generator $M_g$, 
Relation Extractor $M_e$.
\Ensure $Y_{s} \cap Y_{u} = \emptyset$.
\State $M_{g, finetune} \gets Train(M_g, D_s)$
\State $M_{e, finetune} \gets Train(M_e, D_s)$
\State $D_{synthetic} \gets Generate(M_{g, finetune}, Y_u)$
\State $M_{e, final} \gets Train(M_{e, finetune}, D_{synthetic})$
\State $\hat{T_{u}} \gets Predict(M_{e, final}, S_u)$
\\ \Return Extracted Triplets $\hat{T_{u}}$
\end{algorithmic}
\end{algorithm}


\begin{figure}[!t]
\centering
\includegraphics[width=1.0\columnwidth]{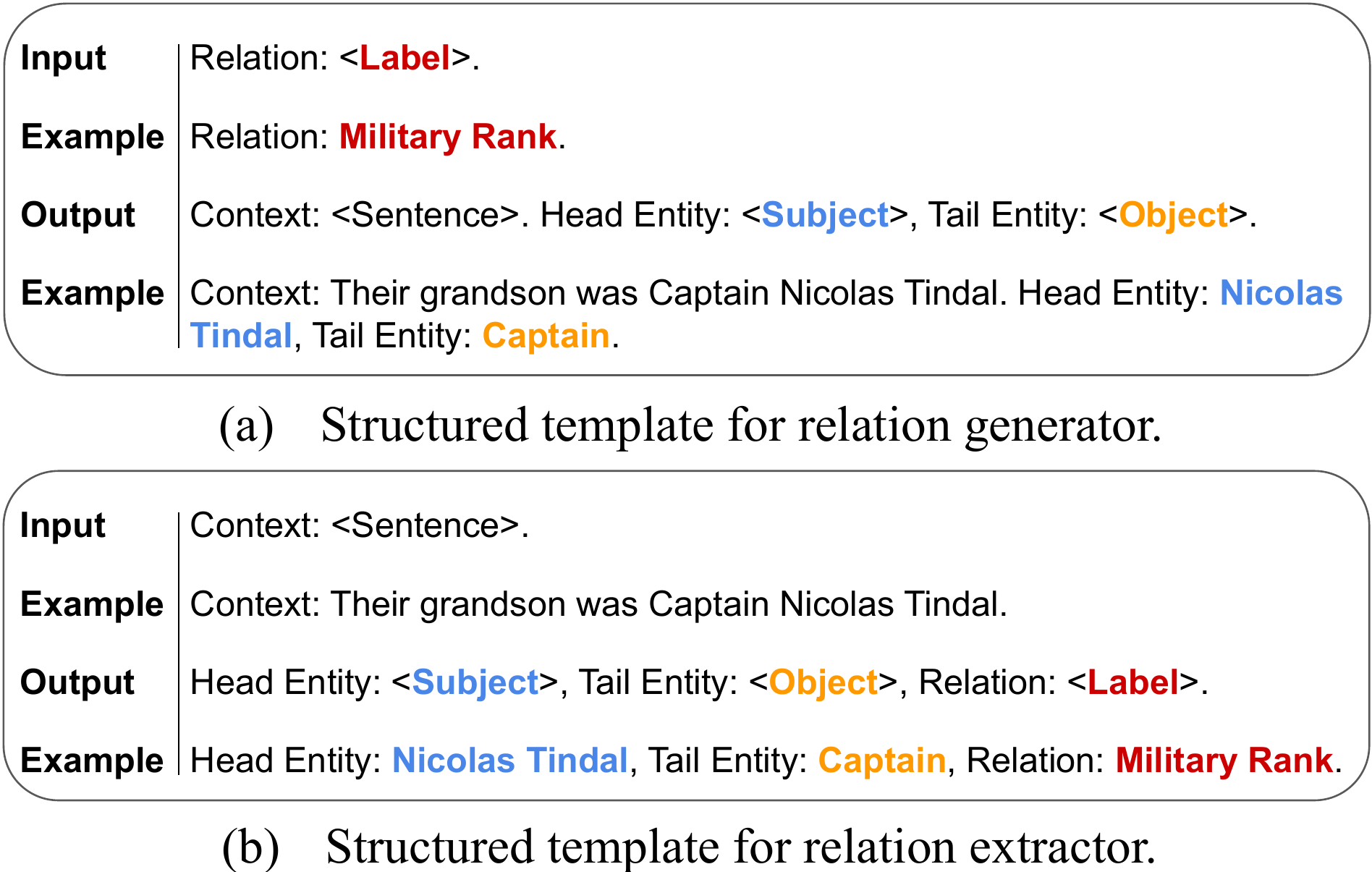}
\caption{
RelationPrompt structured templates.
The head entities, tail entities and relation labels are shown in blue, orange and dark red respectively.
The relation generator (a) takes the relation label as input and outputs the context and entity pair.
The relation extractor (b) takes the sentence as input and outputs the relation triplet which consists of entity pair and relation label. 
}
\label{fig:template}
\end{figure}

\begin{figure*}[!t]
\centering
\includegraphics[width=1.0\textwidth]{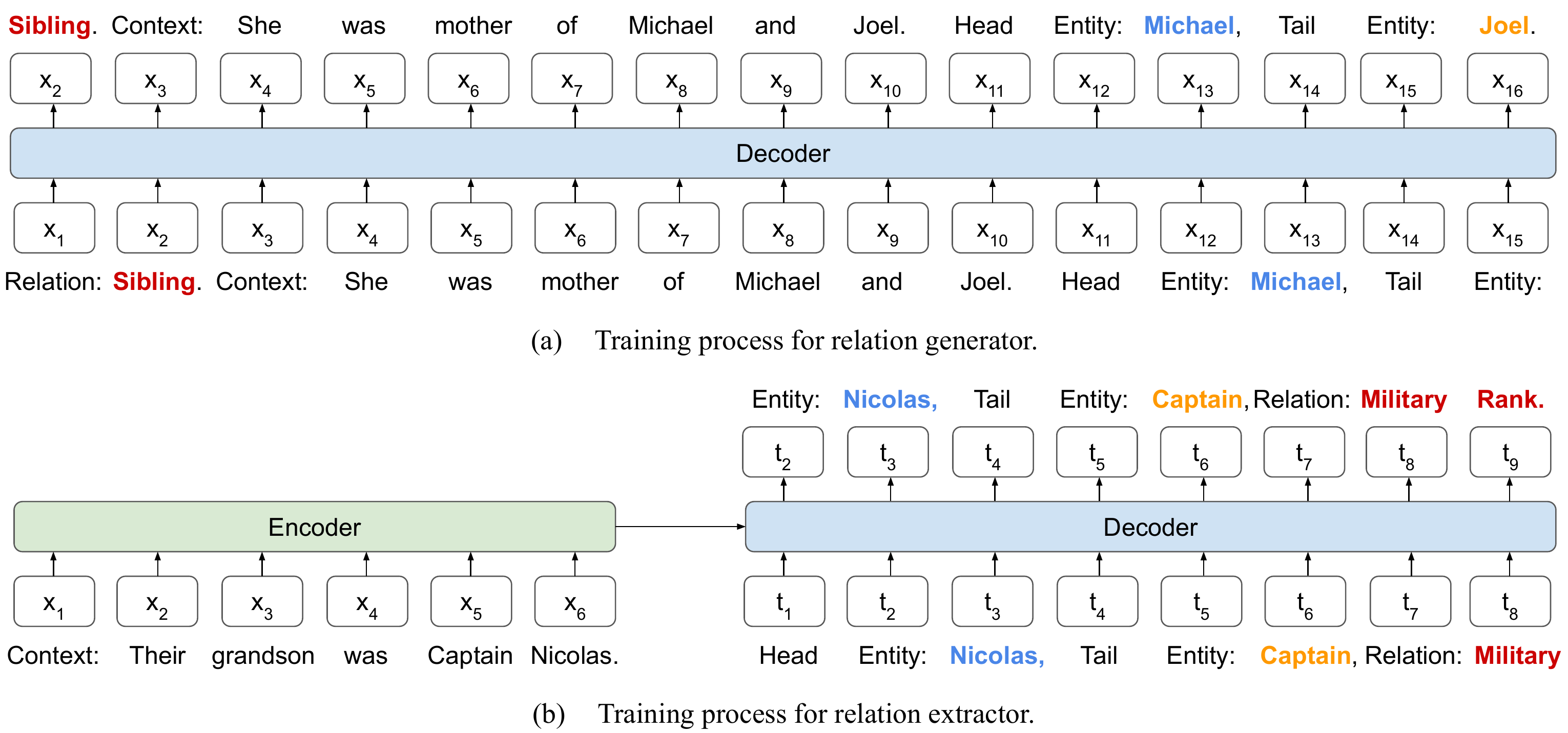}
\caption{
    Model training process. 
    Each head entity, tail entity and relation label is shown in blue, orange and dark red respectively.
    To conserve space, the sentences shown are shortened and punctuation is not separated.
    The relation generator (a) is trained with the standard language modeling objective to condition on the relation label and generate the sentence and entity pair.
    The relation extractor (b) is trained with the standard sequence-to-sequence objective to condition on the input sentence and output the relation triplet of entity pair and relation label.
}
\label{fig:training}
\end{figure*}

\subsection{Relation Generator}
Language models are implicitly capable of zero-shot generalization based on their general and large-scale pre-training \cite{radford2019language}.
Furthermore, text generation has been shown to be effectively controllable \cite{keskar2019ctrl}.
Hence, we prompt the language model to generate synthetic samples by conditioning on the target unseen relation labels.
\flag{
As shown in Algorithm \ref{alg:cap}, relation generator $M_g$ is first fine-tuned on samples for the seen dataset $D_{s}$ (line 1) and then prompted by relation labels $Y_u$ to generate the synthetic samples $D_{synthetic}$ (line 3).
}
As shown in Figure \ref{fig:template}a, the relation generator takes as input a structured prompt in the form of ``Relation: $y$'' and outputs a structured output in the form of ``Context: $s$. Head Entity: $e_{head}$, Tail Entity: $e_{tail}$.''. We employ a causal language model as our relation generator to sample the structured sequence in an autoregressive manner.
As shown in \ref{fig:training}a, the model $M_{g}$ is trained using the standard language modeling objective of next-word prediction \cite{bengio2003neural}.
Given each sequence $x = [x_1, x_2, ..., x_n]$, the goal is to learn the conditional generation probability:
\begin{equation}
    p(x) = \prod_{i=1}^{n} p(x_{i} | x_{<i})
\end{equation}
To generate diverse output sequences for each input relation prompt, we use sampling with temperature $t$ \cite{Hinton2015DistillingTK} over the output logits $o$ and vocabulary size $V$ with temperature $tp$:
\begin{equation}
    p(x_i | x_{<i}) = \frac{exp(o_{i} / tp)}{\sum_{j=1}^{|V|} exp(o_{j} / tp)}
\end{equation}
The output sequences are decoded into relation triplets by splitting on the special terms ``Context:'', ``Head Entity:'' and ``Tail Entity:''.
In case of decoding errors where an entity is not found in the generated context, we discard that sample and continue generating until a fixed amount of valid samples is reached.


\subsection{Relation Extractor}
Given the generated samples of unseen relations, we can train a relation extractor model $M_{e}$ to predict the relation triplets in a zero-shot setting.
\flag{
As shown in Algorithm \ref{alg:cap}, relation extractor $M_e$ is first fine-tuned on samples for the seen dataset $D_{s}$ (line 2) and finally tuned on the synthetic samples $D_{synthetic}$ (line 4).
Lastly, $M_e$ is used to predict and extract relation triplets $\hat{T_u}$ from the test sentences $S_u$ (lines 5 and 6).
}
We adopt a sequence-to-sequence learning approach which is flexible and effective for structured prediction tasks \cite{cui2021template, paolini2020structured}.
As shown in Figure \ref{fig:template}b, the relation extractor takes as input a structured prompt containing the sentence $s$ in the form of ``Context: $s$''. 
It then generates a structured output sequence containing a single pair of entities $e_{head}$ and $e_{tail}$ satisfying the relation $y$, in the form of ``Head Entity: $e_{head}$, Tail Entity: $e_{tail}$, Relation: $y$''.
As shown in Figure \ref{fig:training}b, we use a standard sequence-to-sequence objective \cite{lewis2020bart} for training and greedy decoding for generation.
To predict a single relation triplet in a given sentence $s$, we can generate the model outputs without any initial decoder input, as seen in Figure \ref{fig:decoding}a. 
In case of invalid entity or relation, we treat it as null prediction for that sample.
On the other hand, prediction for ZeroRC is easily supported by providing the entity information as the initial decoder input.
As shown in Figure \ref{fig:decoding}b, the model takes ``Context: $s$, Head Entity: $e_{head}$, Tail Entity: $e_{tail}$, Relation:'' as decoder input to generate ``$y$'' as output.
Hence, our method naturally encompasses both ZeroRTE and ZeroRC as this change affects model prediction and not training.
 

\subsection{Extracting Multiple triplets using Triplet Search Decoding}
We further propose a generation decoding method in order to improve the zero-shot extraction performance on sentences which contain multiple triplets.
For the RelationPrompt generation of synthetic data, each sample is limited to contain a single relation triplet.
Hence, conventional models for triplet extraction most likely cannot perform well with our framework for multi-triplet ZeroRTE as they normally assume that the training samples may contain multiple triplets per sentence.
The inference method of multi-turn question answering \cite{li2019entity} may mitigate this issue, but cannot scale easily to unseen relations as it relies on hand-crafted question templates which are specific to certain relation and entity types.
Hence, we propose Triplet Search Decoding which improves multi-triplet ZeroRTE for the relation extractor.

\begin{figure}[!t]
\centering
\includegraphics[width=1.0\columnwidth]{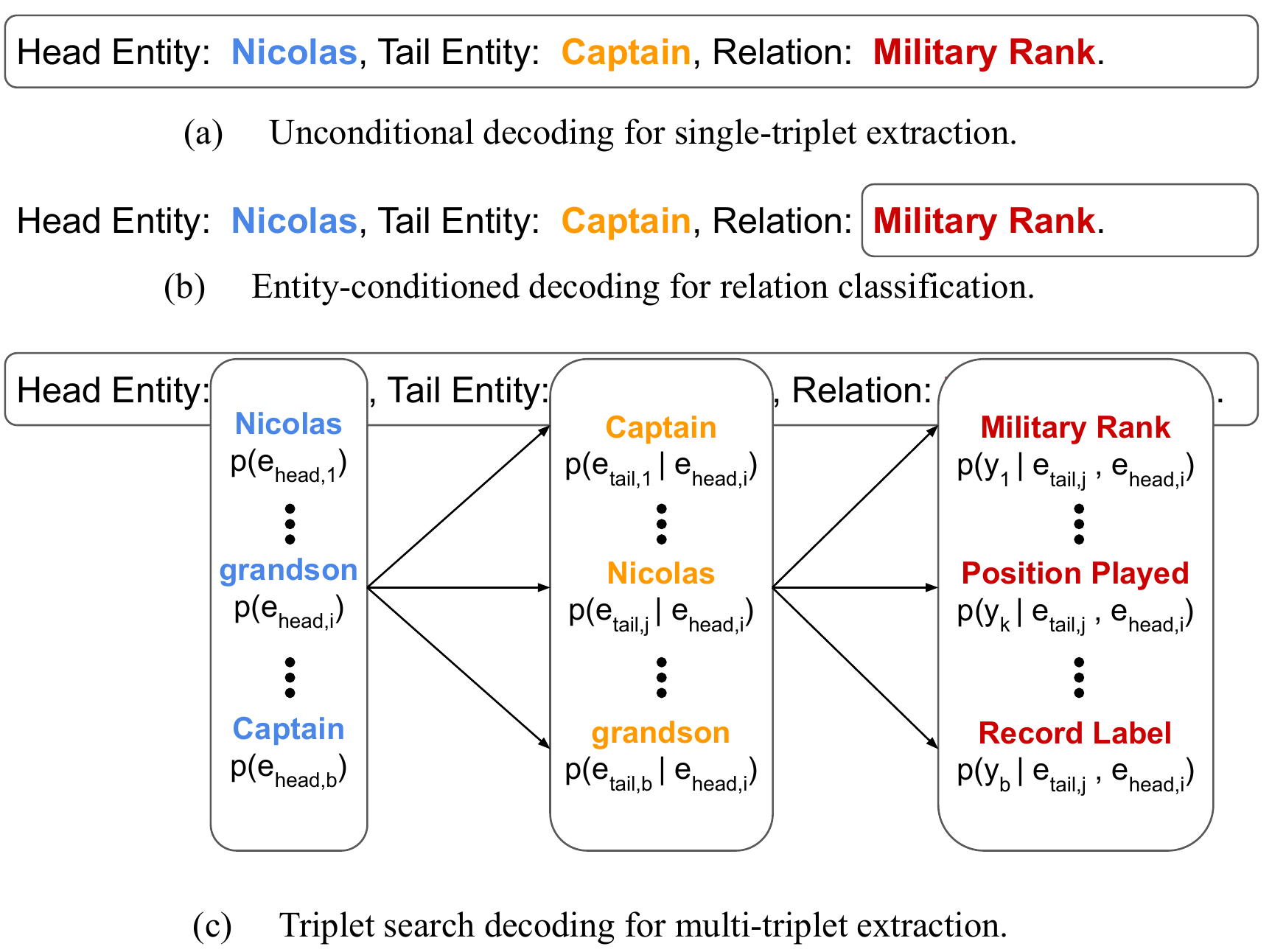}
\caption{
Comparison of generation decoding methods with our proposed Triplet Search Decoding. 
The head entities, tail entities and relation labels are shown in blue, orange and dark red respectively.
Unconditional decoding (a) can be used to predict one relation triplet in each sentence for ZeroRTE. 
Entity-conditioned decoding (b) can be used to predict only the relation label between the given entity pair for ZeroRC.
Our proposed triplet search decoding (c) can be used to predict multiple triplets in each sentence for ZeroRTE.
}
\label{fig:decoding}
\end{figure}

Given the relation extractor which takes a sentence as input and generates output sequences in an autoregressive fashion, greedy decoding as in Figure \ref{fig:decoding}a can output a single sequence. 
However, Triplet Search Decoding as shown in Figure \ref{fig:decoding}c can output multiple sequences that each correspond to a different candidate relation triplet. 
We then apply a likelihood threshold to filter the final output sequences.
The core concept is enumerating multiple output sequences during generation by considering multiple candidates for the head entity, tail entity and relation label respectively.
Starting from the special sub-sequence ``Head Entity:'', it follows from our template in Figure \ref{fig:training}b that the next generated token should be the first token of the head entity, such as ``Nicolas''. 
For the $i^{th}$ possible first token of the head entity, we denote the softmax probability as $p(e_{head,i})$.
We only consider the probability of the first token as it can mostly determine the following generated tokens of the entity \cite{Zhao2021Calibrate}. 
Instead of greedily decoding the entire sequence, we branch the generation into $b$ sequences based on the tokens with the top $b$ highest $p(e_{head,i})$.
Thereafter, the sequence is greedily decoded until the special sub-sequence ``Tail Entity:'' is generated. 
The following token will then be the first token of the tail entity, such as ``Captain''.
The $j^{th}$ tail entity first token probability is denoted as $p(e_{tail,j}|e_{head,i})$.
Hence, the generation is branched such that for each head entity, there will be another $b$ sequences based on the tokens with the top $b$ highest $p(e_{tail,j}|e_{head,i})$.
Thereafter, the sequence is greedily decoded until the special sub-sequence ``Relation:'' is generated. 
The next generated token will be the first token of the relation label, such as ``Military'' in ``Military Rank''.
The $k^{th}$ relation first token probability is denoted as $p(y_{k}|e_{head,i},e_{tail,j})$.
We branch the generation such that for each pair of head entity and tail entity, there will be another $b$ sequences based on the tokens with the top $b$ highest $p(y_{k}|e_{head,i},e_{tail,j})$.
For each sequence, the generation proceeds greedily until the end token is reached, and the overall inference probability is aggregated as:
\begin{equation}
\begin{split}
p(triplet_{i,j,k}) & = p(e_{head,i}, e_{tail,j}, y_{k}) \\
& = p(y_{k}|e_{head,i},e_{tail,j}) \\
& \;\;\;\; \cdot p(e_{tail,j}|e_{head,i}) \\
& \;\;\;\; \cdot p(e_{head,i})
\end{split}
\end{equation}
We note that the conditional assumption does not directly consider the other context tokens as they consist of the special sub-sequences which are fixed as part of our generation template.
The input sentence $s$ is also not included in the formulation as it is the same when considering multiple output triplets for one sample.
At this point, there will be $b^3$ sequences, each corresponding to a different candidate relation triplet.
To filter the final output sequences, we use a probability threshold over that is tuned on the validation $F_1$ metric, with hyperparameter details in Section \ref{sec:implementation}.
Compared to previous generative extraction methods \cite{paolini2020structured, nayak2020effective}, Triplet Search Decoding allows the probability $p(triplet_{i,j,k})$ of each output triplet to be directly calculated and hence control the number of output triplets using the threshold. 
Compared to other decoding methods such as beam search, Triplet Search Decoding leverages the specific relation triplet structure in our structured text templates.
Hence, it can ensure that each output sequence corresponds to a different relation triplet. 
Furthermore, Triplet Search Decoding is more interpretable than existing generative methods for triplet extraction as it can directly provide the prediction probability for each triplet.
More importantly for ZeroRTE, this decoding process allows the relation extractor to naturally predict multiple triplets at test time despite training on synthetic samples which have a single triplet each.

\section{Experiments}
\subsection{Datasets}
\label{sec:exp_data}

We use the following two datasets for our experiments.
FewRel \cite{han2018fewrel} was hand-annotated for few-shot relation extraction, but we made it suitable for the zero-shot setting after data splitting into disjoint relation label sets for training, validation and testing.
Wiki-ZSL \cite{chen2021zs} is constructed through distant supervision over Wikipedia articles and the Wikidata knowledge base.
The dataset statistics are shown in Table \ref{tab:data}.
To partition the data into seen and unseen label sets, we follow the same process as \citet{chen2021zs} to be consistent.
For each dataset, a fixed number of labels are randomly selected as unseen labels while the remaining labels are treated as seen labels during training.
To study the performance of methods under different settings of unseen label set size $m$, we use $m \in \{5, 10, 15\}$ in our experiments.
In order to reduce the effect of experimental noise, the label selection process is repeated for five different random seeds to produce different data folds.
For each data fold, the test set consists of the sentences containing unseen labels.
Five validation labels from the seen labels are used to select sentences for early stopping and hyperparameter tuning. 
The remaining sentences are treated as the train set.
Hence, the zero-shot setting ensures that train, validation and test sentences belong to disjoint label sets.

\begin{table}[!t]
    \centering
    \resizebox{1\columnwidth}{!}{
    \begin{tabular}{llcccc}
    \toprule
    &
    & Samples & Entities & Relations & Sentence Length \\
    \midrule
    & Wiki-ZSL
    & 94,383 & 77,623 & 113 & 24.85 \\
    & FewRel
    & 56,000 & 72,954 & 80 & 24.95 \\

  \bottomrule
    \end{tabular}
    }
    \caption{Dataset statistics. ``Sentence Length'' refers to the average number of words in each sentence.} 
    \label{tab:data}
\end{table}

\begin{table*}[!t]
    \centering
    \resizebox{1.0\textwidth}{!}{%
    \begin{tabular}{llcccccccc}
    \toprule
    \multirow{3}{*}{\textbf{Unseen Labels}}
    & \multirow{3}{*}{\textbf{Model}}  
    & \multicolumn{2}{c}{Single Triplet}
    & \multicolumn{6}{c}{Multi Triplet}
    \\ 
    \cmidrule(lr){3-4} 
    \cmidrule(lr){5-10} 
    & 
    & \multicolumn{1}{c}{\textbf{Wiki-ZSL}} 
    & \multicolumn{1}{c}{\textbf{FewRel}} 
    & \multicolumn{3}{c}{ \textbf{Wiki-ZSL}}  
    & \multicolumn{3}{c}{  \textbf{FewRel}}    
    \\ 
    \cmidrule(lr){3-3} 
    \cmidrule(lr){4-4} 
    \cmidrule(lr){5-7} 
    \cmidrule(lr){8-10}
    && $Acc.$ & $Acc.$& $P.$ & $R.$ & $F_1$ & $P.$ & $R.$ & $F_1$ \\
    
    \midrule
    \multirow{3}{*}{\textbf{m=5}}
    & TableSequence \cite{wang2020two} 
    & 14.47 & 11.82 & \textbf{43.68} & 3.51 & 6.29 & 15.23 & 1.91 & 3.40 \\
    & \textbf{NoGen}
    & 9.05 & 11.49 & 15.58 & \textbf{43.23} & 22.26 & 9.45 & \textbf{36.74} & 14.57 \\
    & \textbf{RelationPrompt}
    & \textbf{16.64} & \textbf{22.27} & 29.11 & 31.00 & \textbf{30.01} & \textbf{20.80} & 24.32 & \textbf{22.34} \\

    \midrule
    \multirow{3}{*}{\textbf{m=10}}
    & TableSequence \cite{wang2020two} 
    & 9.61 & 12.54 & \textbf{45.31} & 3.57 & 6.4 & \textbf{28.93} & 3.60 & 6.37 \\
    & \textbf{NoGen}
    & 7.10 & 12.40 & 9.63 & \textbf{45.01} & 15.70 & 6.40 & \textbf{41.70} & 11.02 \\
    & \textbf{RelationPrompt}
    & \textbf{16.48} & \textbf{23.18} & 30.20 & 32.31 & \textbf{31.19} & 21.59 & 28.68 & \textbf{24.61} \\

    \midrule
    \multirow{3}{*}{\textbf{m=15}}
    & TableSequence \cite{wang2020two} 
    & 9.20 & 11.65 & \textbf{44.43} & 3.53 & 6.39 & \textbf{19.03} & 1.99 & 3.48 \\
    & \textbf{NoGen}
    & 6.61 & 10.93 & 7.25 & \textbf{44.68} & 12.34 & 4.61 & \textbf{36.39} & 8.15 \\
    & \textbf{RelationPrompt}
    & \textbf{16.16} & \textbf{18.97} & 26.19 & 32.12 & \textbf{28.85} & 17.73 & 23.20 & \textbf{20.08} \\

  \bottomrule
    \end{tabular}
    }
    \caption{Results for Zero-Shot Relation Triplet Extraction (ZeroRTE).} 
    \label{tab:main_results}
\end{table*}




\subsection{Experimental Settings}
For the relation generator, we fine-tune the pre-trained GPT-2 \cite{radford2019language} which has 124M parameters. 
For the relation extractor, we fine-tune the pre-trained BART \cite{lewis2020bart} which has 140M parameters.
In both cases, the fine-tuning is performed on the training set for up to five epochs and early stopping is based on the validation loss. 
The learning rate is 3e-5 with linear warm up for the first 20\% of training steps and batch size is set to 128.
During the training process, we use the AdamW optimizer \cite{loshchilov2018decoupled}.
The relation generator is used to generate synthetic samples based on the validation and test set label names. 
A fixed amount of sentences will be generated for each relation.
The relation extractor is fine-tuned again on the synthetic relation sentences and then used for evaluation on the test set.\footnote{See Appendix \ref{sec:implementation} for more implementation details.}

To perform evaluation for ZeroRTE, we evaluate the triplet extraction results separately for sentences containing single triplets and multiple triplets.
To evaluate multiple triplet extraction, we use the Micro $F_{1}$ metric which is standard in structured prediction tasks \cite{paolini2020structured} and report the precision (P.) and recall (R.).
Evaluating single triplet extraction involves only one possible triplet for each sentence, hence the metric used is Accuracy (Acc.).
We evaluate on ZeroRC using the Macro $F_{1}$ metric to be consistent with \citet{chen2021zs}.
Table \ref{tab:main_results} and \ref{tab:zerorc} report the average results across five data folds as detailed in Section \ref{sec:exp_data}.

\subsection{Baseline Methods}
\paragraph{ZeroRTE}
As ZeroRTE is a new task setting, we provide two baseline methods for comparison with our RelationPrompt method. 
Firstly, our relation extractor can be made to perform ZeroRTE without fine-tuning on synthetic samples as it is trained to extract triplets
on the sentences of the seen relation set.
At prediction time, we constrain the generated labels to be selected from the target label names by masking the generated token probabilities. 
We denote this model as 
\textbf{``NoGen''} 
to indicate that it does not use generated synthetic samples for training.
Secondly, we use an existing triplet extraction model known as \textbf{TableSequence} \cite{wang2020two}. 
As it is normally unable to perform ZeroRTE, we provide supervision using synthetic samples from our relation generator.

\paragraph{ZeroRC}
There are three main categories of competing methods for ZeroRC.
Firstly, \textbf{R-BERT} \cite{wu2019enriching} is a relation classification model but can be adapted to the zero-shot setting by using the sentence representations to perform nearest neighbor search over label embeddings.
Next, \textbf{CIM} \cite{rocktaschel2015reasoning} is an entailment-based method which takes the sentence and each possible relation as input to perform binary classification whether the label matches the sentence semantically.
Lastly, \textbf{ZS-BERT} \cite{chen2021zs} generates sentence representations that are conditioned on the provided entity pair information, and performs nearest neighbor search over embeddings of the candidate relation descriptions.

\begin{table}[!t]
    \centering
    \resizebox{1\columnwidth}{!}{%
    \begin{tabular}{llcccccc}
    \toprule
    \textbf{Unseen}
    & \multirow{2}{*}{\textbf{Model}}  
    & \multicolumn{3}{c}{Wiki-ZSL}
    & \multicolumn{3}{c}{FewRel}
    \\ 
    \cmidrule(lr){3-5} 
    \cmidrule(lr){6-8} 
    \textbf{Labels} && $P.$ & $R.$ & $F_1$ & $P.$ & $R.$ & $F_1$ \\
    \midrule
    
    \multirow{5}{*}{\textbf{m=5}}
    & R-BERT
    & 39.22 & 43.27 & 41.15 & 42.19 & 48.61 & 45.17 \\
    & CIM 
    & 49.63 & 48.81 & 49.22 & 58.05 & 61.92 & 59.92 \\
    & ZS-BERT 
    & \textbf{71.54} & 72.39 & 71.96 & 76.96 & 78.86 & 77.90 \\
    & \textbf{NoGen}
    & 51.78 & 46.76 & 48.93 & 72.36 & 58.61 & 64.57 \\
    & \textbf{RelationPrompt}
    & 70.66 & \textbf{83.75} & \textbf{76.63} & \textbf{90.15} & \textbf{88.50} & \textbf{89.30} \\
    
    \midrule
    \multirow{5}{*}{\textbf{m=10}}
    & R-BERT
    & 26.18 & 29.69 & 27.82 & 25.52 & 33.02 & 28.20 \\
    & CIM 
    & 46.54 & 47.90 & 45.57 & 47.39 & 49.11 & 48.23 \\
    & ZS-BERT 
    & 60.51 & 60.98 & 60.74 & 56.92 & 57.59 & 57.25 \\
    & \textbf{NoGen}
    & 54.87 & 36.52 & 43.80 & 66.47 & 48.28 & 55.61 \\
    & \textbf{RelationPrompt}
    & \textbf{68.51} & \textbf{74.76} & \textbf{71.50} & \textbf{80.33} & \textbf{79.62} & \textbf{79.96} \\
    
    \midrule    
    \multirow{5}{*}{\textbf{m=15}}
    & R-BERT
    & 17.31 & 18.82 & 18.03 & 16.95 & 19.37 & 18.08 \\
    & CIM 
    & 29.17 & 30.58 & 29.86 & 31.83 & 33.06 & 32.43 \\
    & ZS-BERT 
    & 34.12 & 34.38 & 34.25 & 35.54 & 38.19 & 36.82 \\
    & \textbf{NoGen}
    & 54.45 & 29.43 & 37.45 & 66.49 & 40.05 & 49.38 \\
    & \textbf{RelationPrompt}
    & \textbf{63.69} & \textbf{67.93} & \textbf{65.74} & \textbf{74.33} & \textbf{72.51} & \textbf{73.40} \\
    
  \bottomrule
    \end{tabular}
    }
    \caption{Zero-Shot Relation Classification (ZeroRC).} 
    \label{tab:zerorc}
\end{table}

    
    

\subsection{Experimental Results}

\paragraph{Triplet Extraction}
We compare RelationPrompt with the baselines on ZeroRTE for Wiki-ZSL and FewRel datasets in Table \ref{tab:main_results}.
In both single-triplet and multi-triplet evaluation, our method consistently outperforms the baseline methods in terms of Accuracy and $F_{1}$ metrics respectively.
Although we do not observe a consistent advantage in precision and recall scores for multi-triplet extraction, the baseline methods cannot achieve a balanced precision-recall ratio, leading to poor overall $F_{1}$ results.
The results difference between NoGen and RelationPrompt also indicate that using the synthetic samples from the relation generator is critical, as the $F_{1}$ score can be improved by more than two times in some cases.
This also suggests that the relation generator can produce reasonable-quality synthetic sentences as training data for the downstream relation extractor.
We also observe that the choice of relation extractor for ZeroRTE is not trivial, as the third-party TableSequence \cite{wang2020two} has significantly worse performance when compared to RelationPrompt, especially for multi-triplet extraction.
Although the TableSequence model is able to perform multi-triplet extraction by design, it assumes that the training data may contain multi-triplet sentences, whereas our synthetic data is limited to single triplet samples.
On the other hand, our proposed relation extractor and decoding method effectively overcomes this challenge by naturally enumerating and ranking multiple triplets at inference time.

\paragraph{Relation Classification} 
RelationPrompt naturally supports
the ZeroRC task without additional training by providing the entity pair information in the prompt.
In Table \ref{tab:zerorc}, we observe consistent improvements compared to the prior state-of-the-art method ZS-BERT \cite{chen2021zs}.
Notably, our method is able to maintain a relatively high classification $F_{1}$ performance when the unseen label set size $m$ increases, whereas ZS-BERT shows a sharper drop in performance.
The trend suggests that RelationPrompt is able to scale better to larger unseen label sets, which is more important for open-domain applications.
This advantage may further indicate that our method can leverage the semantic information of relation labels more effectively through the token-level conditional generation and extraction stages. 
On the other hand, ZS-BERT relies on sequence-level representations which can only preserve the high-level label semantics.

\begin{figure*}[!t]
\centering
\includegraphics[width=0.9\textwidth]{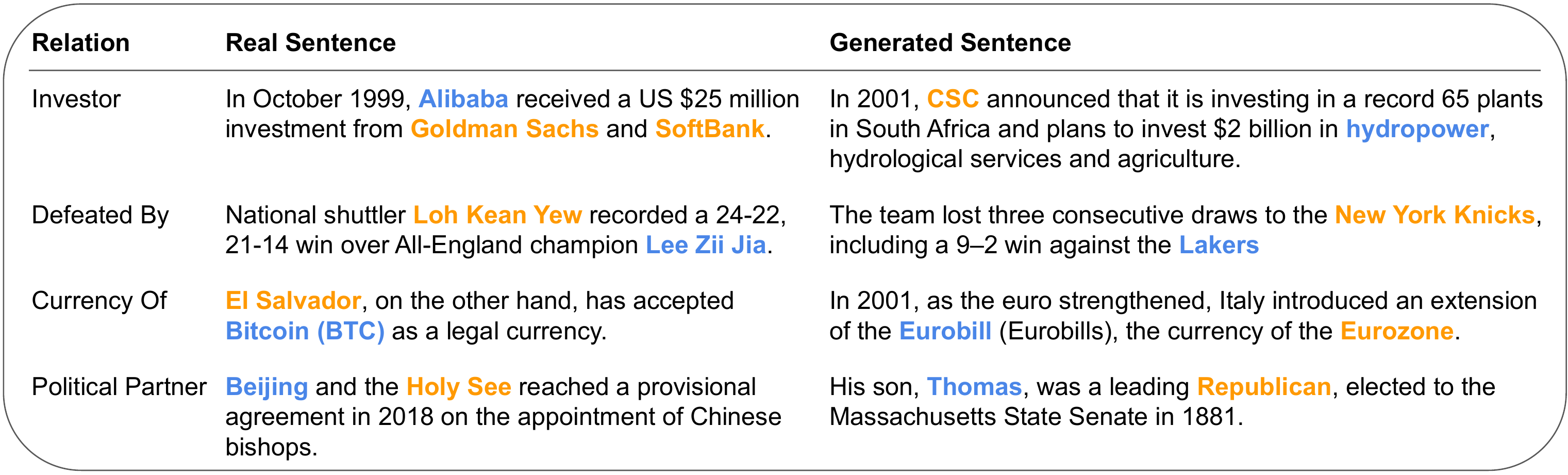}
\caption{
Case study between real and generated samples for relations in the wild.
The head and tail entities are shown in blue and orange respectively.
}
\label{fig:case_study}
\end{figure*}

\begin{table}[t!]
	\centering
	\resizebox{0.99\linewidth}{!}{
		\begin{tabular}{lcc}
			\toprule
			\textbf{Model} & $F_1$ & $\Delta F_1$ \\
			\midrule
			Full Method  
			& \textbf{28.41} & \\
			~~~~$-$ Triplet Search Decoding & 14.53 & -13.88 \\
			~~~~$-$ Extractor Fine-Tuning (Seen Relations) & 13.57 & -14.84 \\
			\bottomrule
		\end{tabular}
	}
	\caption{Ablation results for multi-triplet ZeroRTE. 
	}
	\label{tab:ablation}
\end{table}

\begin{figure}
\centering
\resizebox{0.95\linewidth}{!}{
\begin{tikzpicture}
\pgfplotsset{width = 6cm, height = 3cm}
    \begin{axis}[
        ymax=30,
        ymin=22,
        ylabel={$F_1$ (\%)},
        xlabel={Generated Samples Per Label},
        label style={font=\fontsize{7}{1}\selectfont},
        xtick = {1,2,3,4,5},
        xticklabels = {125, 250, 500, 1000, 2000},
        xticklabel style = {font=\fontsize{7}{1}\selectfont},
        yticklabel style = {font=\fontsize{7}{1}\selectfont},
        xtick pos = left,
        ytick pos = left,
        ymajorgrids = true,
        grid style=dashed,
    ]
    \addplot [mark=square, mark size=1.2pt, color=orange] plot coordinates {
    (1, 22.95) (2, 28.41) (3, 28.01) (4, 27.47) (5, 26.17)};
    \end{axis}
\end{tikzpicture}
}
\caption{Effect of generated data size on ZeroRTE.}
\label{fig:gen_data_size}
\end{figure}
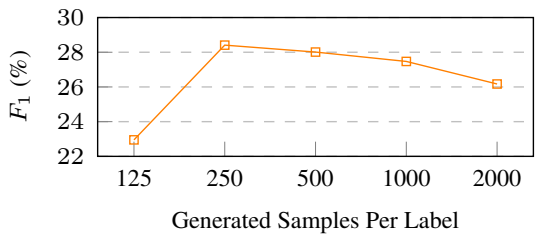

\section{Analysis}

\subsection{Ablation Study}
We conduct an ablation study to examine the performance of our decoding method and task-specific fine-tuning on the seen relation set for multi-triplet ZeroRTE, and the results are shown in Table \ref{tab:ablation}. 
The comparison is conducted on the Wiki-ZSL validation set with 10 unseen labels.
The large performance gap shows that Triplet Search Decoding is critical for multi-triplet ZeroRTE, and suggests that the enumeration and ranking of relation triplet candidates are of sufficiently high quality.
Secondly, we observe a significant drop in performance when the relation extractor is not fine-tuned on seen relation samples from the train set before the final tuning on generated synthetic samples for unseen labels.
This case suggests that the initial fine-tuning on sentences for seen relations is useful for learning the general task of relation triplet extraction.
The learned representations can then be further fine-tuned on the synthetic samples to adapt specifically for the unseen relations to achieve optimal results.



\subsection{Effect of Generated Data Size}
We further study how the number of generated synthetic samples effects the multi-triplet ZeroRTE performance.
The evaluation is based on Wiki-ZSL validation set with 10 unseen labels, and the results are shown in Figure \ref{fig:gen_data_size}. 
Increasing the amount from 125 to 250 samples per label improves $F_{1}$ score.
However, further increasing the generated size up to 2000 does not improve the final performance.
This indicates that although the synthetic data is beneficial for ZeroRTE, excessive amounts can lead to over-fitting due to noise.
We further analyze the generation diversity in Appendix \ref{sec:further_analysis}.

\subsection{Qualitative Analysis}
\label{sec:case}

To assess how the relation data generator generalizes to relations in the wild, we present several samples of real and generated samples in Figure \ref{fig:case_study}.
The relation labels and real sentences were collected from factual articles.
Given the relations ``Investor'', ``Defeated By'' and ``Currency Of'', the generator is able to determine the correct semantic meaning of the relations and compose reasonable sentences.
In most cases, the generated head and tail entity pairings can match the given relations and have a similar context to the real sentences. 
However, in the last case for relation ``Political Partner'', the generated entity pair does not match the relation meaning despite being grounded in a political context.
Instead, the generated sentence expresses a relationship that is closer to ``Political Party''.
This suggests that a future area of improvement could be to match the generated head and tail entity more closely to the given relation.


\section{Related Work}
\paragraph{Zero-Shot Relation Extraction}
Zero-shot relation extraction was previously framed as a slot-filling task and solved by reading comprehension methods \cite{levy2017zero}.
However, their approach requires manual template design for each relation label, which cannot scale well to new relation types.
Another approach to zero-shot relation extraction is the formulation into an entailment task \cite{obamuyide2018zero}, which is not constrained to a fixed relation label space.
Instead, the entailment approach determines if arbitrary pairs of sentences and relation labels are compatible.
However, it is designed for sentence classification and cannot be applied to ZeroRTE.

\paragraph{Data Augmentation}
A popular method for improving model performance in supervised low-resource tasks is data augmentation. 
Simple heuristics such as token manipulation \cite{kobayashi2018contextual} were initially developed, new methods in language modeling improved the quality of augmented samples \cite{xie2020unsupervised, wei2019eda}.
Although there are data augmentation methods that can be applied to structured tasks such as named entity recognition \cite{ding-etal-2020-daga} and relation extraction \cite{papanikolaou2020dare, lee2021neural}, they require existing training samples and cannot be easily adapted to zero-shot tasks.

\paragraph{Knowledge Retrieval}
RelationPrompt also leverages the knowledge stored in language models \cite{roberts2020much} to compose relation samples that are grounded in realistic contexts.
To ensure that the generated samples are factually accurate, the language model requires strong knowledge retrieval capabilities \cite{petroni2019language}.

\paragraph{Language Model Prompts}
Prompting-based methods have shown promise as a new paradigm for zero-shot or few-shot inference in natural language processing \cite{liu2021pre}.
Another advantage is the potential to adapt very large language models \cite{reynolds2021prompt} to new tasks without relatively expensive fine-tuning.
Concurrent works \cite{Meng2022GeneratingTD, Ye2022ZeroGenEZ} also show that language models can generate synthetic training data.
However, such methods have not yet proven effective for more complex tasks such as triplet extraction.

\paragraph{Structured Prediction}
RelationPrompt generates synthetic data for relation triplet extraction, which is a structured prediction task.
Hence, it can be widely applicable to other structured prediction tasks such as named entity recognition \cite{aly-etal-2021-leveraging}, event extraction \cite{huang-etal-2018-zero} or aspect sentiment triplet extraction \cite{xu-etal-2021-learning}.

\section{Conclusions and Future Work}
In this work, we introduce the task setting of Zero-Shot Relation Triplet Extraction (ZeroRTE) to overcome fundamental limitations in previous task settings and encourage further research in low-resource relation extraction. 
To solve ZeroRTE, we propose RelationPrompt and show that language models can effectively generate synthetic training data through relation label prompts to output structured texts.
To overcome the limitation for extracting multiple relation triplets in a sentence, we propose the Triplet Search Decoding method which is effective and interpretable.
Results show that our method surpasses prior ZeroRC methods as well as strong baselines on ZeroRTE, setting the bar for future work.
As mentioned in Section \ref{sec:case}, a future direction for improvement could be to ensure that the generated entity spans are more compatible with the semantics of the relation in the language model prompt.

\bibliographystyle{acl_natbib}
\bibliography{custom}

\clearpage
\newpage

\appendix
\section{Appendix}
\label{sec:appendix}

\begin{figure}[t!]
\centering
\includegraphics[width=1.0\columnwidth]{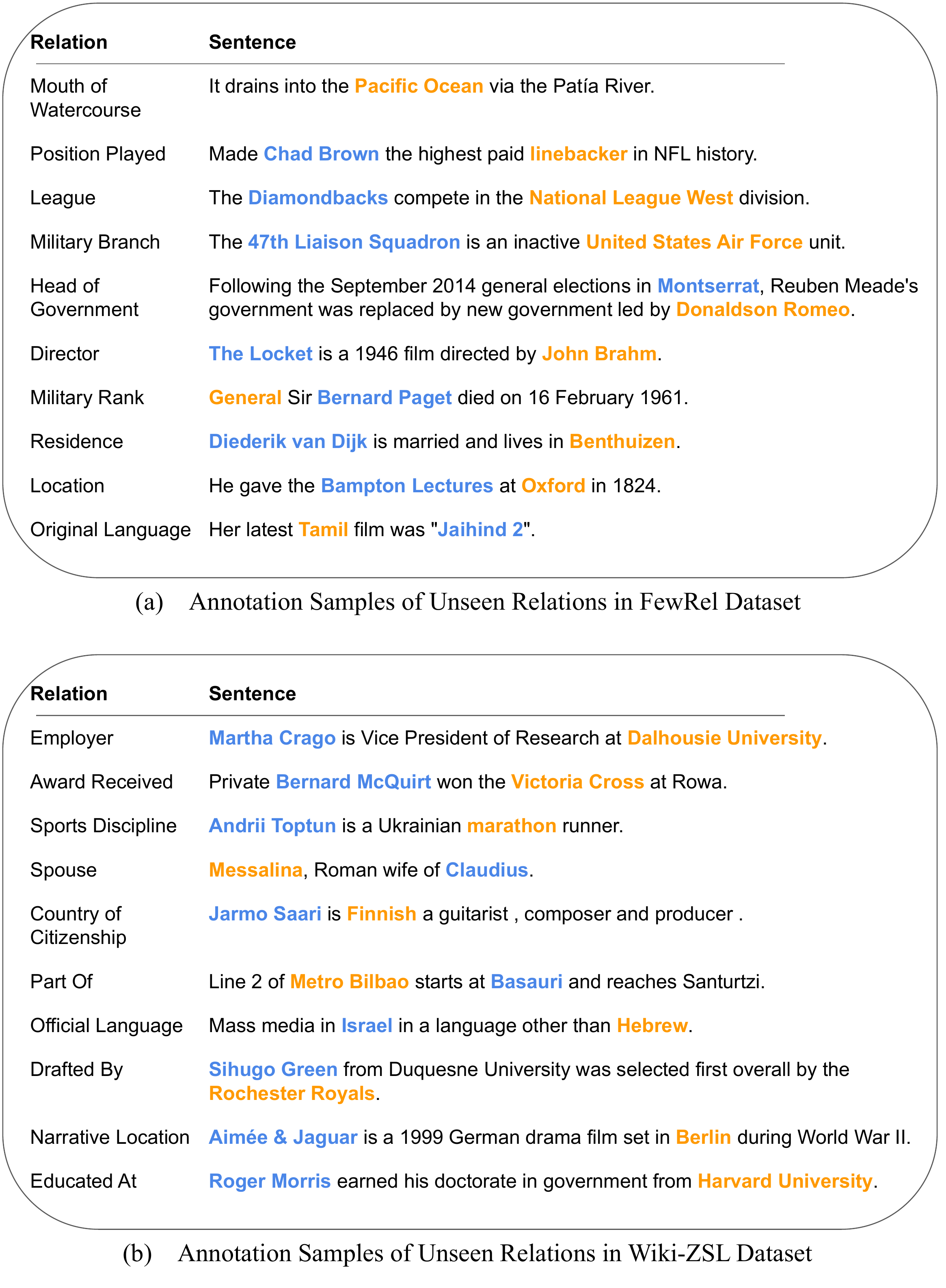}
\caption{
Additional sentence samples from the datasets.
The head and tail entities are shown in blue and orange, respectively.
}
\label{fig:more_examples}
\end{figure}

\begin{figure}[t!]
\centering
\includegraphics[width=1.0\columnwidth]{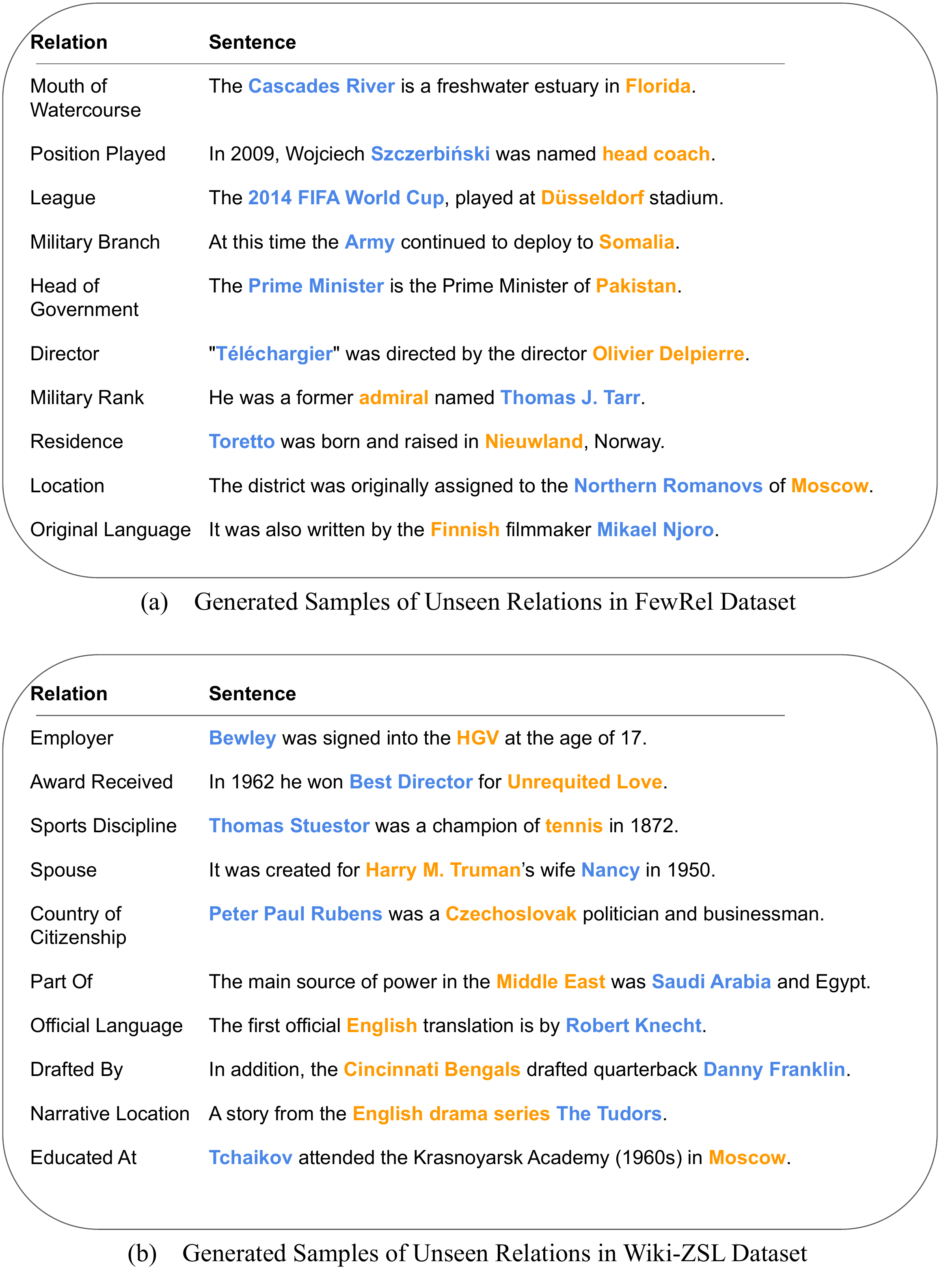}
\caption{
Additional synthetic samples from the generated outputs.
The head and tail entities are shown in blue and orange, respectively.
}
\label{fig:more_synthetic}
\end{figure}

\subsection{Additional Data Samples}
\label{sec:more_samples}

\paragraph{Dataset Samples}
To further illustrate the datasets used, we show test samples in Figure \ref{fig:more_examples}.
The samples are taken from the FewRel (a) and Wiki-ZSL (b) test sets respectively with 10 unseen relation labels.

\paragraph{Synthetic Samples}
To further examine the output of the relation generator, we show test samples in Figure \ref{fig:more_synthetic}.
The samples are generated from the FewRel (a) and Wiki-ZSL (b) test set labels respectively with 10 unseen relation labels.

\subsection{Implementation Details}
\label{sec:implementation}

\paragraph{Generating Structured Texts}
We use the relation generator model to generate synthetic sentences in an autoregressive fashion.
To convert the structured text outputs to relation triplet samples, we perform simple string processing on the output templates shown in Figure \ref{fig:training}a to separate the structured content from the natural text.
In case of a small amount of conversion errors, we continue to generate samples until the amount of sentences generated per label is reached.
For the relation extractor model, we perform a similar processing on the output templates in Figure \ref{fig:training}b to extract the predicted relation triplets.
However, in case of processing errors, we do not continue generation and instead treat it as a prediction failure for that input sample.

\paragraph{Hyperparameters}
We show more detailed hyperparameters used in Table \ref{tab:params}. 
We run a grid search on the Wiki-ZSL validation set with 10 unseen labels for multi-triplet ZeroRTE $F_{1}$ metric.
A grid search is used to tune the hyperparameters.
For number of generated samples per label, we consider the values $\{125, 250, 500, 1000, 2000\}$.
To tune the Triplet Search Decoding threshold, we consider fifty evenly-spaced values from the interval over the minimum and maximum output scores of all candidate triplets on the validation set.
Due to computational constraints, we consider the number of branches to consider at each stage a fixed value, and do not tune it as a hyperparameter.

\paragraph{Computing Infrastructure}
The experiments are conducted on NVIDIA V100 GPUs, and each experiment is run on a single GPU with 32 GB of memory and mixed precision settings.

\subsection{Further Analysis}
\label{sec:further_analysis}
\paragraph{Generated Sample Diversity}

\begin{table}[t!]
    \centering
    \resizebox{1\columnwidth}{!}{
    \begin{tabular}{llc}
    \toprule
    &
    & Value \\
    \midrule
    & Generator Maximum Sequence Length & 128  \\
    & Generator Sampling Top-K & 50 \\
    & Generator Sampling Temperature & 1.0 \\
    & Extractor Maximum Input Length & 128  \\
    & Extractor Maximum Output Length & 128  \\
    & Training Dropout Probability & 0.1 \\
    & Generated Samples Per Label & 250 \\
    & Triplet Search Decoding Top-N Branches & 4 \\
    & Triplet Search Decoding Threshold & -0.9906 \\

  \bottomrule
    \end{tabular}
    }
    \caption{Additional hyperparameters.} 
    \label{tab:params}
\end{table}

Our method for ZeroRTE heavily depends on the quality of the generated data.
Hence, we compare the diversity of real and synthetic data samples.
Concretely, we measure the number of unique words and entities present in the texts.
We used the Wiki-ZSL validation set sentences with five unique labels and generate an equal amount of synthetic sentences for comparison.
Table \ref{tab:diversity} shows that the diversity of unique entities is actually greater for the generated sentences. 
However, the generated sentences have lower diversity of overall unique words.
This may be explained by the fact that entity names tend to be unique, and the generator language model has seen a vast number of unique entity names during the large-scale pre-training.
On the other hand, the total unique words are mostly determined by the non-entity words.
By using prompts to condition the generation of sentences specifically for unseen relation labels, this may constrain the diversity of contextual information in the output sentences.

\begin{table}[t!]
    \centering
    \resizebox{1\columnwidth}{!}{
    \begin{tabular}{llccc}
    \toprule
    &
    & Samples & Unique Entities & Unique Words \\
    \midrule
    & Real Data
    & 3461 & 3090 & 14736 \\
    & Generated Data
    & 3461 & 4949 & 10558 \\

  \bottomrule
    \end{tabular}
    }
    \caption{Data diversity comparison.} 
    \label{tab:diversity}
\end{table}

\begin{figure}
\centering
\resizebox{0.95\linewidth}{!}{
\begin{tikzpicture}
\pgfplotsset{width = 6cm, height = 4cm}
    \begin{axis}[
        ybar,
        ymax=60,
        ymin=0,
        ylabel={$F_1$ (\%)},
        xlabel={Relation Label},
        label style={font=\fontsize{7}{1}\selectfont},
        xtick = {1,2,3,4,5,6,7,8,9,10},
        xticklabels = {
            Official Language, 
            Employer, 
            Part Of, 
            Spouse, 
            Narrative Location,
            Award Received,
            Educated At,
            Country of Citizenship,
            Sports Discipline Competed In,
            Drafted By,
        },
        xticklabel style = {font=\fontsize{7}{1}\selectfont, rotate=90,},
        yticklabel style = {font=\fontsize{7}{1}\selectfont},
        xtick pos = left,
        ytick pos = left,
        ymajorgrids = true,
        grid style=dashed,
    ]
    \addplot  coordinates {
    (1, 5.94) 
    (2, 9.31) 
    (3, 9.55) 
    (4, 12.00) 
    (5, 13.59)
    (6, 15.95)
    (7, 25.95)
    (8, 28.17)
    (9, 47.89)
    (10, 54.41)
    };
    \end{axis}
\end{tikzpicture}
}
\caption{Separate evaluation on relation labels.}
\label{fig:eval_by_label}
\end{figure}
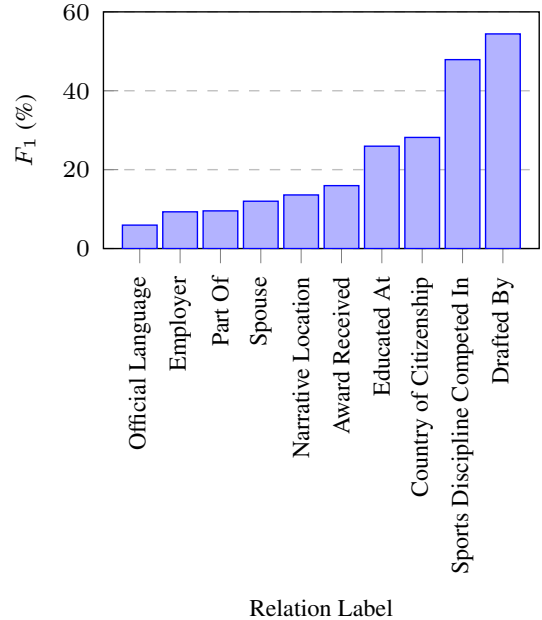


\paragraph{Performance Across Relations}
To study how the performance varies across different relation labels, we evaluate single-triplet ZeroRTE on the Wiki-ZSL test set with 10 unseen labels.
Figure \ref{fig:eval_by_label} shows that the model is able to perform well for relations such as ``Drafted By'' and ``Sports Discipline Competed In''.
However, it performs more poorly for relations such as ``Official Language'' and ``Employer''.
This suggests that RelationPrompt performs best for relations which are highly specific to constrain the output context more effectively.


\end{document}


\maketitle

\section{Methodology}
In order to extract triplets for unseen relation labels, we first formulate ZeroRTE as generating synthetic relation examples of target unseen labels.
Given a set of target relation labels, the goal of RelationPrompt is to prompt language models to generate relation samples which can then be used to supervise any downstream relation extraction model.
Hence, our framework requires two models: a relation data generator to synthesize relation samples, and a relation extractor that will be trained on the synthetic data.
In order to represent the structured relation information that can be processed by language models, we design structured prompt templates.
Although the synthetic relation data is model-agnostic, we design a generative relation extractor and a generation decoding method to overcome the challenge of synthesizing relation samples with multiple triplets.

\subsection{Task Formulation}

The objective of relation triplet extraction is to predict all possible triplets in a given sentence $S$.
Each triplet is defined as 
$(e_{head}, e_{tail}, y)$ where $e_{head}$ and $e_{tail}$ are the head and tail entities respectively, with $y$ as the semantic relationship expressed between them.
For the relation classification objective,
the entity pair $(e_{head}, e_{tail})$ in sentence $S$ is given, and the goal is to predict the relation label $y$. 
In the zero-shot setting, the seen and unseen label sets are denoted as 
$Y_{s} = \{y_{s}^{1}, ..., y_{s}^{n}\}$
and
$Y_{u} = \{y_{u}^{1}, ..., y_{u}^{m}\}$
respectively, where 
$n = |Y_{s}|$
and
$m = |Y_{u}|$
denote the size of seen and unseen label sets respectively.
The seen and unseen label sets are disjoint, i.e., $Y_{s} \cap Y_{u} = \emptyset$.

\begin{figure}[!t]
\centering
\includegraphics[width=1.0\columnwidth]{images/ZeroRTE Template.pdf}
\caption{
Our proposed structured prompt templates.
The head entities, tail entities and relation labels are shown in blue, orange and dark red respectively.
}
\label{fig:template}
\end{figure}

\subsection{Relation Data Generator}
Language models have shown impressive capability to generalize to few-shot and zero-shot settings \cite{radford2018improving, devlin2019bert}. 
This is based on their general and large-scale pre-training which implicitly encompasses a wide range of linguistic tasks \cite{radford2019language}.
Hence, we utilize language models to generate synthetic samples, conditioning on the target unseen relation labels.
Concretely, as shown in Figure \ref{fig:template}a, the generator takes as input a structured prompt in the form of ``Relation: $y$'' and generates a structured output in the form of ``Context: $S$. Head Entity: $e_{head}$, Tail Entity: $e_{tail}$.''. We employ a causal language model $M_{g}$ as our generator to sample the structured sequence in an auto-regressive manner and for simplicity, we denote the process as $M_{g}(y) \rightarrow (S, e_{head}, e_{tail})$. 
We use a standard language modeling objective to train the model over structured text sequences, as shown in Figure \ref{fig:training}a.



\begin{figure*}[!t]
\centering
\includegraphics[width=0.9\textwidth]{images/ZeroRTE Training.pdf}
\caption{
    Model training process. 
    The head entities, tail entities and relation labels are shown in blue, orange and dark red respectively.
    To conserve space, the sentences shown are shortened and punctuation is not separated.
}
\label{fig:training}
\end{figure*}

\begin{figure}[!t]
\centering
\includegraphics[width=1.0\columnwidth]{images/ZeroRTE Decoding.pdf}
\caption{
Relation decoding methods. The head entities, tail entities and relation labels are shown in blue, orange and dark red respectively.
}
\label{fig:decoding}
\end{figure}

\subsection{Generative Relation Extractor}
Given the generated samples of unseen relations, we can train an extractor model which takes a sentence as input and returns a relation triplet as output.
Recent studies show that a wide range of linguistic tasks can be reframed as sequence-to-sequence objectives, such as named entity recognition \cite{cui2021template} and question answering \cite{khashabi2020unifiedqa}.
Hence, we adopt a sequence-to-sequence learning approach for the extractor model $M_{e}$.
Concretely, as shown in Figure \ref{fig:template}b, the relation extractor takes as input a structured prompt containing the sentence $S$ in the form of ``Context: $S$''. 
Consequently, it generates a structured output sequence which contains a single pair of entities $e_{head}$ and $e_{tail}$ satisfying the relation $y$, in the form of ``Head Entity: $e_{head}$, Tail Entity: $e_{tail}$, Relation: $y$''.
For simplicity, we denote the process as $M_{e}(S) \rightarrow (e_{head}, e_{tail}, y)$. 
As shown in Figure \ref{fig:training}b, the extractor model $M_{e}$ is trained using a standard sequence-to-sequence objective over the structured text sequences.
To extract a relation triplet from a given sentence, we can decode the model outputs unconditionally, as seen in Figure \ref{fig:decoding}a.

Note that the above structured template for extraction is designed in the way such that the relation classification task can be performed without changing the model. Instead of generating the output sequence unconditionally as in Figure \ref{fig:decoding}a, for relation classification, we condition the output relation label on the provided head entity and tail entity pair, as shown in Figure \ref{fig:decoding}b.
Specifically, the generator takes ``Context: $S$. Head Entity: $e_{head}$, Tail Entity: $e_{tail}$. Relation:'' as input, and generates ``$y$'' as output.
As this change only affects model prediction and not model training, 
our method naturally encompasses both ZeroRTE and ZeroRC.
 

\subsection{Triplet Search Decoding}
We further propose a generation decoding method in order to improve the extraction performance on sentences which contain multiple triplets and rank the possible triplets in a more transparent manner. 
Currently, the generator $M_{g}$ and extractor $M_{e}$ are designed to generate and extract a single triplet for each 
\flag{sentence}
respectively. 
However, it is common for sentences to express multiple semantic relationships between entities, thus containing multiple relation triplets. 
\flag{
Although sequence-to-sequence models can be applied to triplet extraction tasks \cite{paolini2020structured, nayak2020effective}, they cannot be easily adapted to multi-triplet extraction for the zero-shot setting as they assume that the training data may contain multi-triplet sentences in order to perform inference, whereas our synthetic relation samples are limited to a single triplet each.
The inference method of multi-turn question answering \cite{li2019entity} may mitigate this issue, but it cannot scale easily to unseen relations as it relies on hand-crafted question templates which are specific to certain relation and entity types.
Furthermore, it is not clear how to assign the prediction likelihood to each possible triplet or control how many triplets to generate.
Hence, we propose Triplet Search Decoding which enables multi-triplet extraction in a transparent and controllable manner for sequence-to-sequence models.
}

Given the relation extractor which takes a sentence as input, the goal is to \flag{generate structured texts} from the decoder such that we can enumerate all possible triplets and perform ranking over the top-scoring triplet candidates.
Concretely, our template in Figure \ref{fig:template}b is designed to facilitate this as the head entity, tail entity and relation label are spans of tokens which form a linear sequence, separated by the special phrases.
For instance, Figure \ref{fig:decoding}c shows that the sequence generation begins with the phrase ``Head Entity:'', followed by the first token of the head entity.
Instead of decoding the first token of the span through methods such as greedy or beam search, we branch the generation into multiple sequences by selecting different beginning tokens for the entity such as ``Nicolas'', ``grandson'', ``Captain'', etc.
Although the generated span may require multiple tokens, span generation is mostly determined by the first token \cite{Zhao2021Calibrate}. 
Therefore, we only perform this branching in three stages: the first token of the head entity, tail entity and relation label, respectively.
\flag{
Each possible path represents a triplet candidate.
With reference to Figure \ref{fig:decoding}c, the likelihood of each branch is represented by the generation probability of the respective initial token.
Consequently, the overall likelihood of the path is represented by the aggregated likelihood of each branching stage.
As the full enumeration of all paths is not computationally practical, we only consider the top four branches at each stage for our experiments.
Hence, a total of 64 triplet candidates are considered.
To select the most suitable triplet candidates, we use a score threshold that is tuned on the validation set, and each triplet candidate with overall path likelihood higher than the threshold is predicted as an output relation triplet.
}

\bibliographystyle{acl_natbib}
\bibliography{custom}